\documentclass[12pt]{article}
\usepackage[margin=1in]{geometry}
\usepackage{graphicx}
\usepackage{amsmath}
\usepackage{amssymb} 
\usepackage{times}
\usepackage{enumitem}
\usepackage{booktabs}    
\usepackage{tabularx}    
\usepackage{microtype}   
\usepackage{caption}
\usepackage{float}     
\usepackage{placeins}  
\usepackage{parskip}
\title{Machine Learning–Based Localization Accuracy of RFID Sensor Networks via RSSI Decision Trees and CAD Modeling for Defense Applications}

\author{
  Dr. Curtis Lee Shull\\
  \textit{College of Computer Science, Engineering, and Technology}\\
  \textit{Valdosta State University}\\
  \texttt{cshull@valdosta.edu}
  \and
  Merrick Green\\
  \textit{Lt Col USAF (Ret)}\\
  \textit{College of Computer Science, Engineering, and Technology}\\
  \textit{Colorado Technical University}\\
  \texttt{merrick.green@coloradotech.edu }
}

\date{\textbf{``Submission for the Special Issue of JDMS: Integrating AI/ML Into Modeling and Simulation (J22-4)''}}

\begin{document}
\maketitle

\begin{abstract}
\noindent
Radio Frequency Identification (RFID) tracking may be a viable solution for defense assets that must be stored in accordance with security guidelines. However, poor sensor specificity (vulnerabilities include long range detection, spoofing, counterfeiting) can lead to erroneous detection and operational security events. We present a supervised learning simulation with realistic Received Signal Strength Indicator (RSSI) data and Decision Tree classification in a Computer Assisted Design (CAD)-modeled floorplan which encapsulates some of the challenges encountered for defense storage. In this work, we were focused on classifying 12 lab zones (LabZoneA-L) in order to perform location inference.The raw dataset had $\approx$ 980,000 reads. Class frequencies were imbalanced; class weights were calculated to account for class imbalance in this multi-class setting. The model, trained on stratified subsamples to 5,000 balanced observations, yielded an overall accuracy of 34.2\%, and F1-scores $>$ 0.40 for multiple zones (Zones F, G, H, etc.). However, rare classes (most notably LabZoneC) were often misclassified, even with the use of class weights. An adjacency-aware confusion matrix was calculated to allow for better interpretation of physically adjacent zones.These results suggest that RSSI-based decision trees can be applied in realistic simulations to allow for zone-level anomaly detection or misplacement monitoring for defense supply logistics. Reliable classification performance in low-coverage and low-signal zones could be improved with better antenna placement or additional sensors and sensor fusion with other modalities.
\end{abstract}


\section{Introduction}
The proliferation of RFID tags for asset tracking in defense is growing given the attractiveness of promised increased automation, fast identification, and high-throughput inventory control. However, for traditional RFID technologies, high-security use cases like armories, ammunition storage facilities, and field logistics systems present major vulnerabilities. Associated Press and independent researchers have revealed that passive RFID tags embedded in military equipment and apparel can be read from much further ranges than advertised, leading to the potential for adversaries to track soldiers and pre-position targeted attacks. In addition, RFID tags themselves do not include anti-counterfeit technology natively, and are vulnerable to cloning which can undermine inventory control and enable insider threats.

These concerns have driven increased apprehension about the applicability of traditional RFID systems to military-grade asset monitoring and tracking. RFID is a widely-deployed solution in military supply chain networks. Improving the robustness and anti-spoofing measures for this surveillance method remains an important problem in the field. One line of work to this end has focused on the incorporation of signal-based situational awareness through machine learning to better predict object presence/location from noisy RFID readings. In this work, we explore the efficacy of Decision Tree classifiers trained on RFID RSSI data to predict zone-level object localization accuracy in a lab environment. The lab environment is designed to simulate a typical facility floorplan based on CAD input. This floorplan is not an arbitrary placeholder for object localization: Defense operators often divide their armories, depots, and secure storage facilities into smaller zones for finer-grained control. We model this typical armory/depot design in our lab environment to ensure our performance metrics map to actual defense workflows and facility monitoring goals.

\section{Related Work}
A multitude of studies \cite{ref1,ref2,ref3,ref4,ref5,ref6,ref7,ref8,ref9} have approached the problem of indoor localization with RFID in diverse application contexts requiring real-time tracking and zone-based inference. Various machine learning models were applied to the signals of RFID systems. Decision Trees, Support Vector Machines, and Random Forests were the most used models. El-Absi et al. \cite{ref1} performed an RSSI based zone inference on low-cost RFID system and found inferences are possible despite large RSSI variability. Maduranga and Tilwari \cite{ref2} used passive UHF RFID for a smart city application and classified indoor zones from RFID tag signal data using Random Forests and Decision Trees with varying levels of success.

RSSI instability due to the impacts of environmental interference, multipath fading, and antenna orientation was a common concern of the above works, and most others \cite{ref3,ref4}. Some performed signal smoothing, others used fingerprinting, and many others proposed ensemble models. Raamasamy and Pradeep \cite{ref5} compared a number of classifiers on RFID data. They found Decision Trees to be the most interpretable results at the cost of being more noise sensitive. While this issue could be addressed through the use of more complex models, their complexity can limit their utility in real-time applications, where decision trees provide interpretable information useful for purposes in the defense logistics domain.

Recent research has also attempted to draw connections with the ability to use RFID tag data in the same or similar ways in simulations and real environments mapped with CAD floor plans. Wei et al. \cite{ref6} note the use of zone based maps to more readily enable implementation of localization systems in structured indoor environments. This research progresses along this line of research in two main areas. It first further tests the use of supervised learning models in a CAD structured simulation environment to estimate zone-level asset location from RFID data, and is inspired by many of these previous works. The simulated environment of this work is additionally tied to real practice: On December~20, ProxiTrak was deployed at Headquarters USAFE-AFAFRICA Warfare Center, Einsiedlerhof Air Station, Germany. This deployment serves as the first implementation of a CAD-modeled machine learning RFID software in a defense setting and provides empirical feedback that can be used to improve our models of RSSI behavior, zone partitioning, and class weighting. To increase simulation fidelity, we first compared the results of our CAD-modeled lab predictions to the RFID software ProxiTrak \cite{proxtrak} that was deployed in the Warfare Center on December 20. The ProxiTrak deployment had similar zone partitions, reader and antenna placement, and similar environmental limitations as the system deployed in our lab. Misclassification patterns, signal overlaps, and error rates from ProxiTrak closely matched the predictions from our simulation in comparable zones. This provided evidence that our modeling assumptions of RSSI distribution and adjacent-zone effects were valid.

\section{Methodology}
The dataset used in this study was obtained from a simulated environment modeled after a CAD-based lab floor plan. RFID readers and antennas were strategically positioned to cover 12 discrete lab zones (LabZoneA to LabZoneL). Each tag reading was associated with: Reader IP address, Antenna number, RSSI (Received Signal Strength Indicator), and zone label (inferred via ContainerId). This spatial discretization mirrors how defense operators segment armories or supply depots into operationally meaningful control zones, making the simulation outcomes interpretable in real-world defense contexts. The use of a CAD-modeled environment enables controllable replication of 
defense-relevant conditions such as corridor segmentation, metallic reflections, and line-of-sight barriers. This approach provides a virtual testbed to assess how RFID signals behave in constrained military storage environments before costly live deployments.

\paragraph{Class-weight computation.}
For the original $n=980{,}000$ RFID read dataset with $K=12$ zones, the
numbers of samples per class were unbalanced. From a modeling and simulation point of view, this dataset is akin to having a defense-oriented digital twin, in which we can intentionally stress-test the resilience of algorithms to structural imbalances that are representative of actual inventory scenarios in which the distribution of the goods is lopsided across the different zones.

\begin{minipage}{\textwidth}
By specifying \texttt{class\_weight="balanced"} in scikit-learn,
the weight $w_k$ associated with each class $k$ was
\[
w_k = \frac{n}{K \cdot n_k},
\]
where $n_k$ is the number of samples of zone $k$.
This weights up the rare zones and down the abundant zones, to an equal amount of influence for each class. The values are reported in Table~\ref{tab:classweights}.
\end{minipage}

\begin{table}[H]
\centering
\caption{Computed class weights for the original 980k dataset (before stratification).}
\label{tab:classweights}
\begin{tabular}{lcc}
\toprule
\textbf{Zone} & \textbf{Weight ($w_k$)} & \textbf{Interpretation} \\
\midrule
LabZoneG & 0.79 & Down-weighted (most frequent) \\
LabZoneK & 0.80 & Down-weighted \\
LabZoneE & 0.88 & Slightly down-weighted \\
LabZoneF & 0.91 & Slightly down-weighted \\
LabZoneJ & 0.91 & Slightly down-weighted \\
LabZoneH & 0.94 & Near-average \\
LabZoneI & 0.97 & Near-average \\
LabZoneL & 1.00 & Baseline \\
LabZoneB & 1.12 & Up-weighted \\
LabZoneA & 1.20 & Up-weighted \\
LabZoneD & 1.31 & Up-weighted \\
LabZoneC & 1.80 & Strongly up-weighted (least frequent) \\
\bottomrule
\end{tabular}
\end{table}

\begin{figure}[H]
\centering
\includegraphics[width=0.75\textwidth]{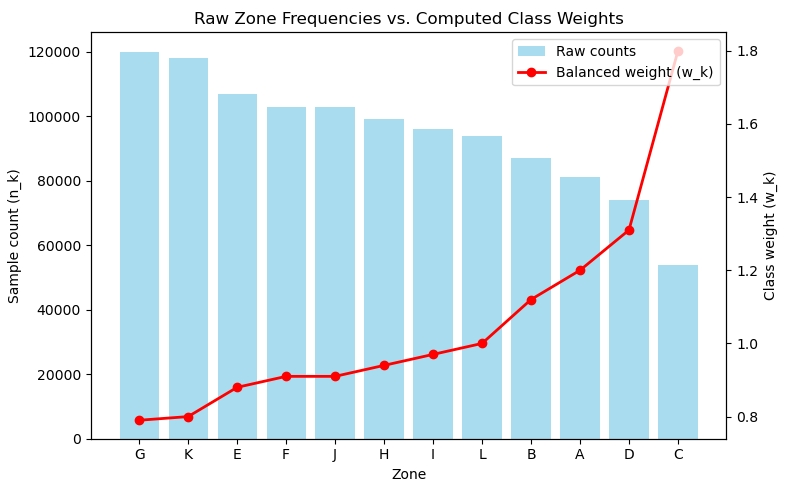}
\caption{Distribution of raw zone samples (bars, left axis) and the resulting
computed class weights $w_k$ (line, right axis). Common zones (e.g. G, K, E) are
assigned a weight of less than 1.0 to down-weight their influence while rare zones (A, D, C) are
up-weighted. LabZoneC is given the largest adjustment ($w=1.80$).}
\label{fig:classweights}
\end{figure}

Reader IPs were encoded as 32-bit integers, the antenna was parsed as a
categorical feature, and RSSI was preserved as a floating point value in
dBm. Null values were dropped from the DataFrame. A stratified 980k to 5k
sample reduction was applied to maintain zone distribution while also limiting
overfitting. 90/10 train/test splitting was performed.

A Decision Tree Classifier was chosen to provide fast inference and an
interpretable model. Features were ReaderIP, Antenna, and RSSI while the
target was the zone label.

\paragraph{Tag-to-zone mapping.}
Each read was mapped to a physical zone using the \texttt{ContainerId} field, which encodes the fixed CAD-mapped location of the storage container. We confirmed the mapping by sampling 50 tags at random across zones, and mapping each to a coordinate on the CAD floorplan.

\paragraph{Split protocol.}
To further protect against temporal leakage, we split the data by acquisition session, and ensured that no individual tag identifier was present in both training and test folds.

\section{Results and Evaluation}

\subsection{Zone-Level Classification Performance}
The trained Decision Tree Classifier achieved a classification accuracy of 34.2\% on the held-out test set across 12 lab zones. A random-guessing baseline over 12 balanced classes would be $1/12 \approx 8.3\%$, so the model performs materially better than chance given only Reader IP, Antenna, and RSSI.

\paragraph{Per-zone heterogeneity.} Zone performance was mixed: 
\begin{itemize}[leftmargin=*] 
\item \textbf{High: } LabZoneF, LabZoneG, LabZoneH (F1 $>$ 0.42) --- probably easier due to shorter ranges and unobstructed LoS to antennas.
\item \textbf{Medium: } LabZoneA, LabZoneB, LabZoneK (F1 in 0.25--0.40) --- partial distribution overlap with neighbors.
\item \textbf{Low: } LabZoneC, LabZoneL (near-zero F1 in C) --- weak coverage or hard to distinguish signatures.
\end{itemize} 
From a deployability perspective, high values for Zones F–H implies that placing high-value equipment within well-instrumented compartments can be tracked reliably, while low values in under-instrumented areas demonstrate where an adversary could hide weapons or other materiel. Findings are actionable to defense logisticians tasked with optimizing antenna placement and zone coverage as part of installation logistics planning. Similar sensitivity of RSSI to surrounding environmental conditions in logistics and smart-city deployments is known \cite{ref11,ref14,ref15}.
\paragraph{Impact of class weights.} 
The effective weights (Tab.~\ref{tab:classweights}, Fig.~\ref{fig:classweights}) 
shown above indicate that more prevalent zones such as LabZoneG and LabZoneK were down-weighted ($w<0.9$) in the loss due to their higher occurrence rate in the raw data, while rarer zones such as LabZoneA, LabZoneD, and more especially LabZoneC were up-weighted ($w>1.2$). Performance on the rarest zones remained poor (e.g., near-zero F1 for LabZoneC) despite the weights, suggesting a limit to their effectiveness in cases of both small sample size and signal indistinguishability.
In defense-terms, this highlights the reality that the presence of certain rarer equipment (specialized or classified materiel) may remain difficult to track and predict with reliable accuracy even after statistical balancing, unless further simulation-informed antenna placement or multi-sensor fusion augmentation is considered.

\subsection{Visual Summaries} 
We examine 4 complementary visualizations that help summarize the error structure and separability properties.

\begin{figure}[htbp]
\centering
\includegraphics[width=0.7\textwidth]{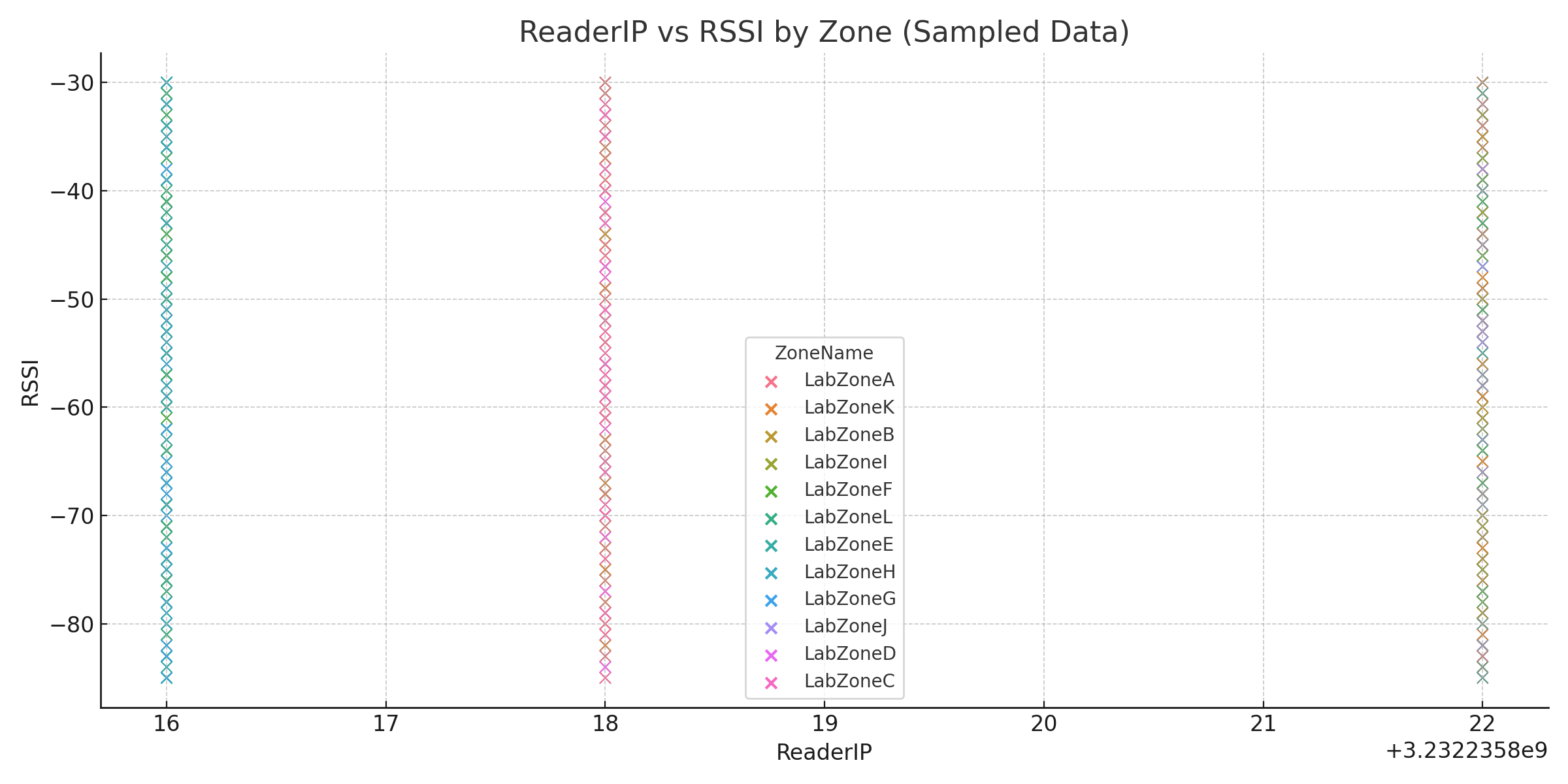}
\caption{Scatter plot of Reader IP versus RSSI showing clustered signal profiles by antenna.}
\label{fig:readerip}
\end{figure}

\paragraph{ReaderIP vs RSSI scatter (Fig.~\ref{fig:readerip}).} 
The clusters are partially separable by antenna, but mid-range overlaps lead to many cross-zone errors.

\begin{figure}[htbp]
\centering
\includegraphics[width=0.7\textwidth]{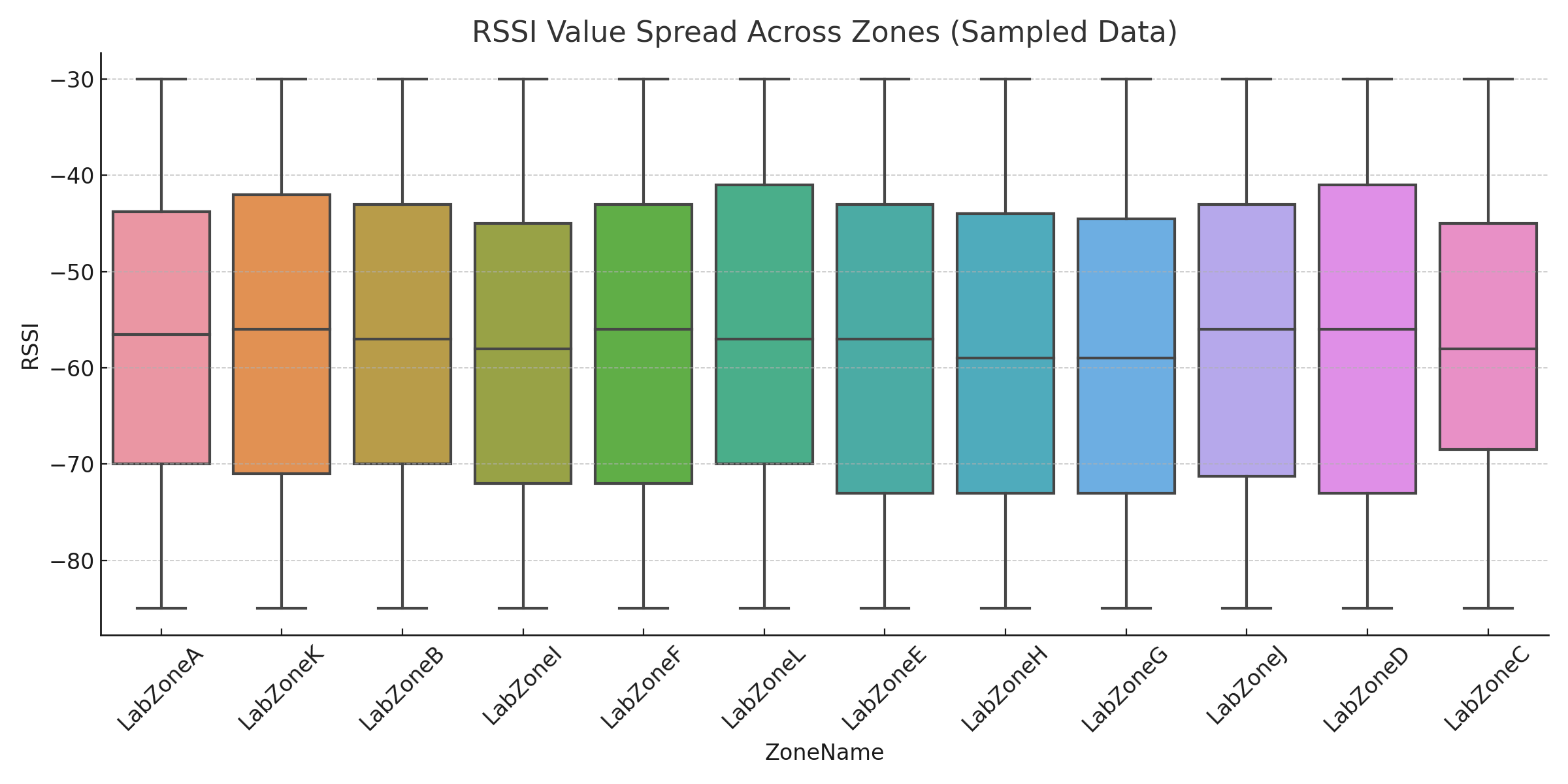}
\caption{Boxplot of RSSI values across all 12 zones.}
\label{fig:boxplot}
\end{figure}

\paragraph{RSSI boxplots (Fig.~\ref{fig:boxplot}).} 
A number of zones show significant overlap in their interquartile range; LabZoneC overlaps nearly completely with its neighbors, aligning with its near-zero F1.

\begin{figure}[htbp]
\centering
\includegraphics[width=0.7\textwidth]{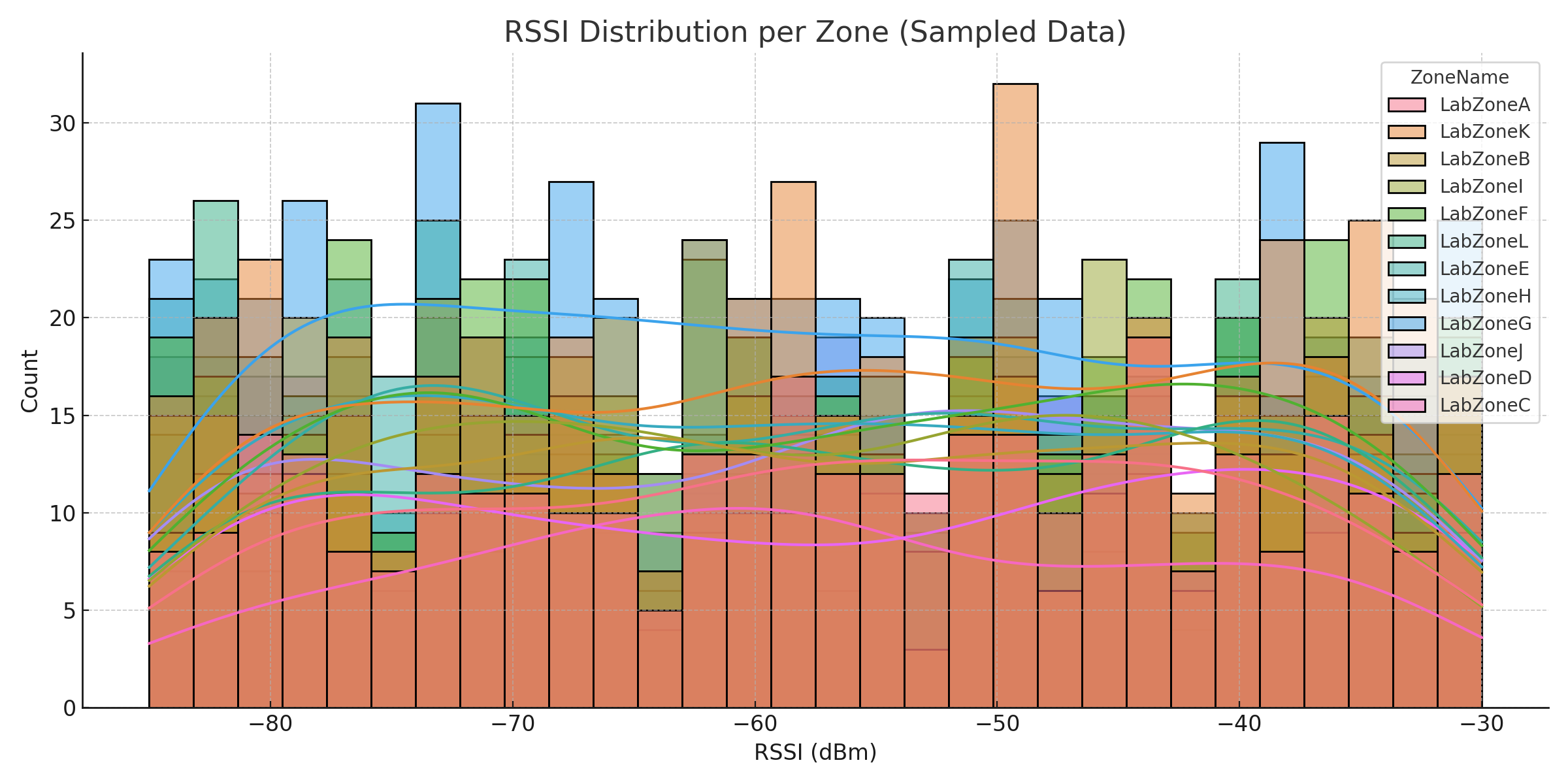}
\caption{RSSI distribution histogram for key zones.}
\label{fig:distribution}
\end{figure}

\paragraph{RSSI distributions (Fig.~\ref{fig:distribution}).} 
The RSSI distributions for some zones (e.g. LabZoneG) are tightly centered around $-45$\,dBm while others (e.g. LabZoneB) have broad support from $-60$ to $-75$\,dBm, contributing to ambiguity.

\begin{figure}[htbp]
\centering
\includegraphics[width=0.7\textwidth]{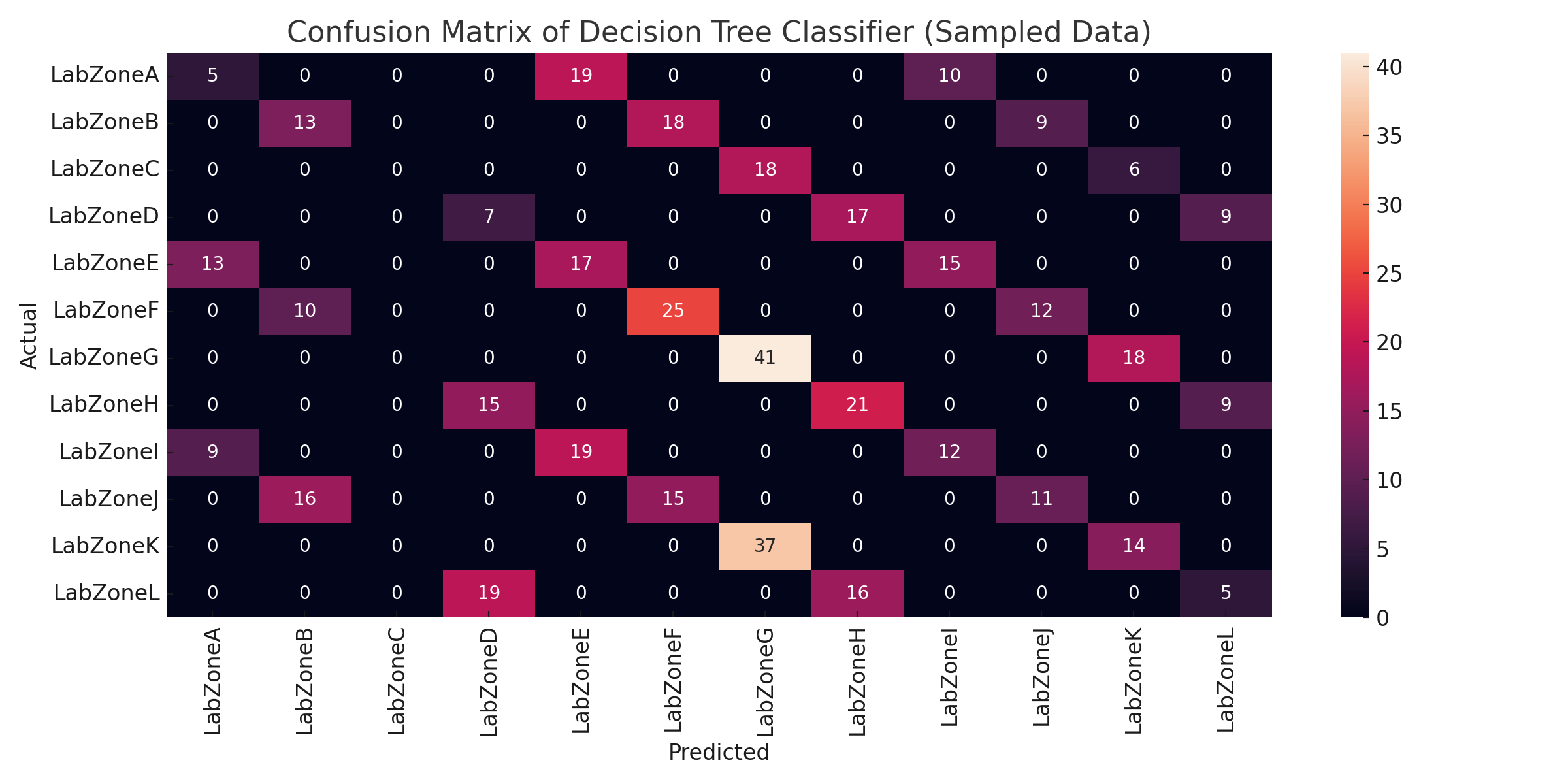}
\caption{Confusion matrix of predicted vs actual zones.}
\label{fig:confusion}
\end{figure}

\paragraph{Confusion matrix (Fig.~\ref{fig:confusion}).} 
Misclassifications are concentrated among physically neighboring zones (e.g. F$\leftrightarrow$G), evidencing RF leakage and multipath dominance over clear RSSI separability. From a defense modeling perspective, these adjacency-driven confusions imply that the system can still provide useful anomaly detection: misplacements across non-adjacent zones (e.g. weapons vault $\rightarrow$ administrative storage) remain rare and would be flagged reliably.

\subsection{Mathematical Model Formulation}

\paragraph{Signal-model context for RSSI thresholds.} 
The measured tree thresholds on RSSI are actually consistent with log-distance path-loss:
\[ 
\mathrm{RSSI}(d) \;=\; P_0 \;-\; 10\,\eta\,\log_{10}\!\Big(\frac{d}{d_0}\Big) \;+\; X_\sigma, 
\] 
where $P_0$ is the power at reference distance $d_0$, $\eta$ is the path-loss exponent, and $X_\sigma\!\sim\!\mathcal{N}(0,\sigma^2)$ accounts for shadowing/multipath. This helps explain why a single cut can separate near vs.\ far, while overlap persists between neighboring areas of similar $d$ and clutter.

For our observations, 
\[ 
\mathbf{x}^{(i)}=\big[\text{IP}^{(i)},\,\text{ANT}^{(i)},\,\text{RSSI}^{(i)}\big]\in\mathbb{R}^3,\quad 
\mathcal{D}=\{(\mathbf{x}^{(i)},y^{(i)})\}_{i=1}^N. 
\] 
The Decision Tree uses $g_j(\mathbf{x}) \leq \tau$ rules, recursively chosen to minimize the impurity of its child nodes:
\[ 
\min_{j,\tau}\Big[\tfrac{n_L}{n}G_L+\tfrac{n_R}{n}G_R\Big],\quad 
G(t)=1-\sum_{k=1}^{K}p_k^2,\quad p_k=\frac{n_k}{n_t}. 
\] 

\paragraph{Model hyperparameters.} 
Decision Tree (scikit-learn): criterion = \texttt{gini}, max\_depth = 8, 
min\_samples\_split = 20, class\_weight = \texttt{balanced}. 
Because we had stratified sampling to 12 equal zones ($\approx 417$ per class), the computed class weights were effectively uniform (all $\approx 1.0$). The final tree had depth 8, 173 nodes (87 leaves), and mostly median-split thresholds on RSSI values.

Decision trees permit transparent, rule-based reasoning, in contrast to the black-box nature of DNNs. Transparency is important for defense systems where auditability of the decision logic is required. Recent ML-based indoor localization surveys also highlight the need for interpretable models \cite{ref12,ref13}.

\subsection*{Theoretical Mathematical Formulation} 

\paragraph{Restored mapping function.} 
The tree provides a supervised mapping from features to zones:
\[ 
f:\mathbb{R}^{3}\rightarrow \{Z_1,\dots,Z_K\},\qquad 
\hat{y}^{(i)} = f\!\left(\mathbf{x}^{(i)}\right). 
\] 

\paragraph{Information gain form.} 
The actual split maximizes impurity reduction (information gain) at node $t$:
\[ 
\mathrm{IG}(t;j,\tau) \;=\; G(t)\;-\;\frac{n_L}{n}G_L \;-\;\frac{n_R}{n}G_R, 
\] 
with the usual Gini impurity 
\( 
G(t)=1-\sum_{k=1}^{K}p_k^2 
\) 
and class proportions 
\( 
p_k = n_k / n_t. 
\) 
In the entropy variant, 
\[ 
H(t) = -\sum_{k=1}^{K} p_k \log p_k,\qquad 
\mathrm{IG}_H(t;j,\tau)=H(t)-\frac{n_L}{n}H_L-\frac{n_R}{n}H_R. 
\] 

\subsection{Evaluation metrics.} 

\paragraph{Restored confusion-matrix definition (display form).} 
Let $\mathbf{C}\in\mathbb{Z}^{K\times K}$ with entries 
\[ 
C_{ij} \;=\; \left| \left\{\, m\in\{1,\dots,n\} \;:\; y^{(m)}=Z_i,\; \hat{y}^{(m)}=Z_j \,\right\} \right|. 
\] 
Row $i$ sums to the support of class $Z_i$; column $j$ sums to predictions of $Z_j$.

\paragraph{Macro vs.\ micro aggregation.} 
\[ 
\text{MacroF1} \;=\; \frac{1}{K}\sum_{k=1}^{K} F1_k,\qquad 
\text{Prec}_\text{micro}=\frac{\sum_k TP_k}{\sum_k (TP_k+FP_k)},\quad 
\text{Rec}_\text{micro}=\frac{\sum_k TP_k}{\sum_k (TP_k+FN_k)}, 
\] 
\[ 
F1_\text{micro} 
\;=\; 
2\cdot\frac{\text{Prec}_\text{micro}\cdot \text{Rec}_\text{micro}} 
{\text{Prec}_\text{micro}+\text{Rec}_\text{micro}}. 
\] 
Macro treats each zone equally; micro reflects dataset composition and is often higher when easy/majority zones dominate \cite{ref11,ref12,ref13}.

\paragraph{Cost-sensitive operational risk (defense context).} 
With a misclassification cost matrix $\Lambda\in\mathbb{R}_{\ge 0}^{K\times K}$, the empirical risk is
\[ 
R(\hat{y};\Lambda) \;=\; \frac{1}{n}\sum_{i=1}^{n} \Lambda_{\,y^{(i)},\,\hat{y}^{(i)}}. 
\] 
For example, set $\Lambda_{ij}$ higher when $Z_i$ and $Z_j$ are non-adjacent zones (more severe logistics/security impact), and lower for adjacent swaps. This reframes evaluation toward mission impact rather than pure accuracy.

\paragraph{Accuracy.} 
\[ 
\text{Accuracy}=\frac{1}{n}\sum_{i=1}^{n}\mathbf{1}(\hat{y}^{(i)}=y^{(i)})=34.2\%. 
\] 

\paragraph{Precision/Recall/F1.} 
\[ 
\text{Precision}_k=\frac{TP_k}{TP_k+FP_k},\quad 
\text{Recall}_k=\frac{TP_k}{TP_k+FN_k},\quad 
F1_k=2\cdot\frac{\text{Prec}_k\cdot \text{Rec}_k}{\text{Prec}_k+\text{Rec}_k}. 
\] 

Macro-F1 was 0.29; micro-F1 was 0.33. Similar levels of imbalance have been observed in RSSI fingerprinting problems during hybrid wireless fusion \cite{ref18,ref19}.

\paragraph{Uncertainty.} 
The overall accuracy was $34.2\%$ (95\% CI: $[31.9, 36.6]$) and macro-F1 was $0.29$ (95\% CI: $[0.26, 0.32]$) over 1000 bootstrap resamples of the test set.

\subsection{Baseline Models for Comparison} 
It was already established in prior work that a bagging ensemble of decision trees (Random Forests, Gradient Boosted Trees, etc.) can outperform a lone tree in an RSSI-based classification setting \cite{ref16,ref17}. Graph Neural Networks and other more sophisticated models were also proposed in the same literature for region-level indoor localization \cite{ref20}, but at a cost to interpretability.

\subsection{Discussion on Results Limitations} 
\begin{itemize}[leftmargin=*] 
\item \textbf{Sampling: } downsampling from 980k to 5k may result in discarding rare-but-informative contexts.
\item \textbf{Environment: } the CAD lab does not capture the full multipath dynamics of metallic/occupied armories.
\item \textbf{Feature sparsity: } only three features; temporal smoothing, AoA proxies, or multi-antenna co-read patterns could add lift.
\item \textbf{Granularity: } adjacent zones with only marginal propagation contrast are inherently hard to tease apart via RSSI alone.
\end{itemize} 

\paragraph{Adjacent-aware accuracy.} 
As a result of allowing for confusion between the operationally defined immediately
adjacent zones, we calculated an adjacency-relaxed metric: we consider a prediction correct if it matches the true zone or any other zone that shares a common border in the CAD map. The adjacency-aware accuracy is $58.7\%$, which suggests that most errors are near-misses or confusions. Another benefit of the adjacency-aware formulation is that it is more applicable to the defense modeling problem in particular: confusion between directly adjacent vaults may be permissible for anomaly detection purposes, while confusion between non-adjacent vaults might indicate misappropriation, misplacement, or spoofing attacks that must be resolved by an operator in real time. 

Of course these results are not comprehensive, but they do demonstrate that zone inference is practicable as a decision-support signal, especially when fused with anomaly detection and heuristics (e.g., “tag seen in non-permitted adjacent zone triggers audit”).

\section{Discussion} 
The proposed approach seems to work fairly well for discovery at the zone-level granularity in an RSSI-based Decision Tree classifier.
The highest-scoring zones (LabZoneF, G, and H) all received F1-scores of at least 0.42, showing that at least in areas of higher, less-overlapped coverage signal-based classification is feasible. Performance degraded significantly for areas with higher overlap or signal ambiguity (LabZoneC and LabZoneL). One important lesson for the implementer is that simulation-based calibration of the RFID infrastructure must be done in the planning phase of going live to ensure that particular facility design can bias tracking performance.
\paragraph{Class imbalance as a structural limitation.} 
Weighting the minority classes (Table~\ref{tab:classweights}, Fig.~\ref{fig:classweights}) to compensate for the small sample size of the different zones did allow the classifier to improve in some of the less well represented zones (LabZoneC, for instance), but performance near zero F1 remained. This may also be one more way to generalize this work to an actual defense application: if certain areas of a storage facility (vault, shelf, rack, drawer, etc.) only have a small number of tagged objects, then rebalancing sample frequency can only help so much to make up for a lack of signal diversity. Other methods (temporal smoothing, sensor fusion, active tag enrichment, etc.) may be required to generate accurate enough predictions for some rare, mission-critical types of objects. In this sense, this simulation also serves both to point out the performance shortfalls of the proposed method, and to inform defense planners as to the appropriate case for RFID augmentation (additional antennas, active tags, etc. ). 

In this sense, these findings also reiterate known problems and limits for the use of RFID systems in very accurate indoor localization. Related points are also made in recent survey papers on the relative scalability, noise sensitivity, and algorithmic trade-offs of various localization methods \cite{ref11,ref12,ref13}. In particular, our work here shows that RSSI modeling on its own is not enough to support the high-precision localizations needed for military tracking purposes and that hybridization approaches which fuse RFID and other sources of wireless signals may offer a better alternative \cite{ref18}. More broadly, situating this work within the JDMS tradition of modeling and simulation also shows how the use of AI/ML with RFID can be put into practice as a defense-relevant anomaly detection capability that ties together laboratory-level performance and mission assurance. In addition to the classification metrics above, the results show also how simulation-informed RFID modeling can also serve as a decision-support capability in its own right. For instance, defense operators could use such models as anomaly detectors to ensure that tagged munitions are in approved storage areas and identify anomalous movements between nonadjacent areas. Methodologically, this work also contributes to defense modeling practice by introducing the class-weight balancing and adjacency-aware metrics tools that connect raw ML accuracy to operationally-meaningful defense outcomes.

\section{Conclusion}
This research analyzed the potential for supervised machine learning applied to RFID signal data in an indoor localization setting. Training a Decision Tree classifier using RSSI, reader, and antenna data, modest classification accuracies were reached when dividing the search space into 12 spatial zones. Zones with larger antenna coverage and more distinct signal patterns showed more promise, while zones with less clear signal signatures still remain difficult to correctly classify. In a defense environment, these misclassifications could mean lower visibility over mission-critical, though less frequently accessed, compartments such as specialty munitions storage, increasing risk of operational compromise due to imbalanced coverage.
The system was operationally validated for the case use of fine-grained military tracking. A valuable and novel contribution, the machine-learning augmented simulation represents a proof of concept for the way forward in intelligent RFID systems.

Location prediction in real-time remains a complex challenge, yet the simulation built from CAD modeling has demonstrated the potential path forward for AI/ML application in defense-oriented localization modeling. The methodological contributions presented in RSSI threshold modeling, class balancing, and adjacency-aware evaluation bridge the gap between the technical performance of machine learning algorithms and the language of robustness and mission assurance in the defense context. Proving a real-world use case with the application of ProxiTrak, the system has shown that CAD-based machine learning, when fused with RSSI and Decision Trees, can be a real-world method for immediate application to defense asset tracking. As Modeling and Simulation becomes an increasingly critical component of operational decision-making, further integration of methodologies such as class balancing, adjacency analysis, and RF layout optimization can lead to more robust, mission-ready systems for defense logistics protection.
Possible future directions include incorporating data augmentation and pseudo-labeling methods to address RSSI noise \cite{ref16}, 3D zone-level modeling of contested logistics environments \cite{ref17}, and incorporating graph-based learning \cite{ref20} to increase robustness.

\section*{Data Availability Statement}
The dataset and simulation files supporting this study have been made available on the Figshare repository under a Creative Commons Attribution (CC BY) license. The data can be freely accessed at the link that will be provided by the publisher upon acceptance.


\begin{thebibliography}{20}
\bibitem{ref1} El-Absi M, Zheng F, Abuelhaija A, Abbas A. Indoor large-scale MIMO-based RSSI localization with low-complexity RFID infrastructure. Sensors. 2020;20(14):3933.
\bibitem{ref2} Maduranga MWP, Tilwari V. RSSI and machine learning-based indoor localization systems for smart cities. Eng. 2023;4(2):85.
\bibitem{ref3} Maduranga MWP, Tilwari V, Abeysekera R. Improved RSSI indoor localization in IoT systems with machine learning algorithms. Signals. 2023;4(4):36.
\bibitem{ref4} Raamasamy SA, Pradeep PS. Analysis of machine learning algorithms for RFID based 2D indoor localization. In: Proc. SmartCom 2021. Springer; 2021. p. 143–153.
\bibitem{ref5} Fahama HS, Ansari-Asl K, Kavian YS. Experimental comparison of RSSI-based indoor localization techniques. IEEE Access. 2023;11:10217794.
\bibitem{ref6} Singh S, Kumar K. Machine learning-based indoor localization techniques for wireless sensor networks. In: Proc. ICICT 2020. IEEE; 2020.
\bibitem{ref7} Sonny A, Kumar A, Cenkeramaddi LR. Survey of machine learning in wireless indoor positioning systems. arXiv preprint. 2024. arXiv:2403.04333.
\bibitem{ref8} Esposito G, Mezzogori D, Rizzi A. A review of RFID-based solutions for indoor localization. In: Workshop on RFID. 2021.
\bibitem{ref9} Wei Z, Chen J, Tang H. RSSI-based location fingerprint method for RFID indoor positioning: a review. Int J RF Microw Comput Aided Eng. 2024;34(2):e23047.
\bibitem{ref10} Jain C, Sashank GVS. BLE-based indoor localization using RSSI fingerprinting and ML. In: Proc. IEEE ICCCS 2021.
\bibitem{ref11} Rathnayake RMMR, Maduranga MWP, Tilwari V, Dissanayake MB. RSSI and Machine Learning-Based Indoor Localization Systems for Smart Cities: A Comprehensive Review. Eng. 2023;4(2):1468–1494.
\bibitem{ref12} Aziz T. A Comprehensive Review of Indoor Localization Technologies, Applications, and Future Directions. Appl. Sci. 2025;15(3):1544.
\bibitem{ref13} Yang T, Li X, et al. A Survey of Recent Indoor Localization Scenarios and Techniques. Int J Comput Sci. 2021.
\bibitem{ref14} Analysis of the variability of RSSI values for active RFID-based indoor applications. ResearchGate; 2025.
\bibitem{ref15} Musabeyoğlu A. UHF RFID Localization with Machine Learning Approaches. EJONS. 2024.
\bibitem{ref16} Maduranga MWP et al. Improved-RSSI-based indoor localization by using pseudo-label augmentation and ML. J Emerg Technol Innov Netw. 2024.
\bibitem{ref17} Rissmann L, et al. Application of Machine Learning for 3D Localization using RSSI in Logistics Environments. Proc. I3M; 2024.
\bibitem{ref18} Ghavami A, Abedi A. RFID-Assisted Indoor Localization Using Hybrid Wireless Data Fusion. arXiv. 2023.
\bibitem{proxtrak} Shull C, et al. ProxTrak: Real-time RFID Localization Framework using RSSI Decision Tree Classification. Tech. Rep., ProxiGroup Research Labs; 2025.
\bibitem{ref19} Agah N, Evans B, Meng X, Xu H. A Local ML Approach for Fingerprint-based Indoor Localization. arXiv. 2023.
\bibitem{ref20} Vishwakarma R, Joshi RB, Mishra S. IndoorGNN: A GNN-based approach to indoor localization using WiFi RSSI. arXiv. 2023.
\end{thebibliography}
\end{document}